\documentclass{IEEEtran}
\IEEEoverridecommandlockouts
\usepackage{cite}
\usepackage{amsmath,amssymb,amsfonts}
\usepackage{algorithmic}
\usepackage{algorithm}
\usepackage{graphicx}
\usepackage{textcomp}
\usepackage{xcolor}

\usepackage{amsthm}
\usepackage{amsmath}
\usepackage{booktabs}
\usepackage{pifont}
\usepackage{makecell}
\usepackage{amsfonts,amssymb}
\usepackage{subfigure}
\usepackage{multirow}
\usepackage[table]{xcolor}

\def\BibTeX{{\rm B\kern-.05em{\sc i\kern-.025em b}\kern-.08em
    T\kern-.1667em\lower.7ex\hbox{E}\kern-.125emX}}
\begin{document}

\title{Copy-Augmented Representation for \\ Structure Invariant Template-Free Retrosynthesis
\\ \thanks{This work was supported by Shanghai Frontiers Science Center of Molecule Intelligent Syntheses.}
}

% \author{\IEEEauthorblockN{1\textsuperscript{st} Jiaxi Zhuang}
% \IEEEauthorblockA{\textit{School of Computer Science and Technology}\\
% \textit{East China Normal University}\\
% Shanghai, China \\
% 51255901063@stu.ecnu.edu.cn}
% \and
% \IEEEauthorblockN{2\textsuperscript{nd} Ying Qian*}
% \IEEEauthorblockA{\textit{School of Computer Science and Technology}\\
% \textit{East China Normal University}\\
% Shanghai, China \\
% yqian@cs.ecnu.edu.cn}
% }

\author{
    \IEEEauthorblockN{
        Jiaxi Zhuang$^{1*}$, \quad Yu Zhang$^{1*}$, \quad Aimin Zhou$^{2\dagger}$, \quad Ying Qian$^{2\dagger}$}\\
    % \vspace{0.2cm}
    \IEEEauthorblockA{
        \textit{$^1$School of Computer Science and Technology, East China Normal University} \\
        \textit{$^2$Shanghai Institute of Artificial Intelligence for Education, East China Normal University} \\
        % \{52295901029, 51285901011\}@stu.ecnu.edu.cn \\
        % \{amzhou, yqian\}@cs.ecnu.edu.cn \\
        $^{*}$Equal contribution; \quad $^{\dagger}$Corresponding author.
    }
}

\maketitle

\begin{abstract}
Retrosynthesis prediction is fundamental to drug discovery and chemical synthesis, requiring the identification of reactants that can produce a target molecule. Current template-free methods struggle to capture the \textit{structural invariance} inherent in chemical reactions, where substantial molecular scaffolds remain unchanged, leading to unnecessarily large search spaces and reduced prediction accuracy. We introduce C-SMILES, a novel molecular representation that decomposes traditional SMILES into element-token pairs with five special tokens, effectively minimizing editing distance between reactants and products. Building upon this representation, we incorporate a copy-augmented mechanism that dynamically determines whether to generate new tokens or preserve unchanged molecular fragments from the product. Our approach integrates SMILES alignment guidance to enhance attention consistency with ground-truth atom mappings, enabling more chemically coherent predictions. Comprehensive evaluation on USPTO-50K and large-scale USPTO-FULL datasets demonstrates significant improvements: 67.2\% top-1 accuracy on USPTO-50K and 50.8\% on USPTO-FULL, with 99.9\% validity in generated molecules. This work establishes a new paradigm for structure-aware molecular generation with direct applications in computational drug discovery.
\end{abstract}


\section{Introduction}

Retrosynthesis prediction represents a fundamental challenge in computational drug discovery and chemical synthesis, requiring the identification of synthetic pathways from target molecules to available starting materials. Originally proposed by Corey \cite{retrofirst}, retrosynthesis has traditionally relied on extensive expertise in organic chemistry and reaction mechanisms. With the rapid advancement of artificial intelligence in chemical sciences, computational retrosynthesis has emerged as a critical tool for accelerating pharmaceutical development and synthetic chemistry \cite{segler2018planning,corey1985computer}.

Recent computational approaches fall into three main categories: template-based methods \cite{retrosim,gln} that rely on predefined reaction templates, semi-template approaches \cite{g2gs,xpert,graphretro} that combine template matching with generative models, and template-free methods that treat retrosynthesis as a sequence-to-sequence translation problem. Among these, template-free methods \cite{scrop,augtrans,retroformer} have gained particular attention due to their scalability and ability to generalize beyond predefined template libraries. These approaches represent molecules using Simplified Molecular Input Line Entry System (SMILES) \cite{smiles} and frame retrosynthesis prediction as a machine translation task, enabling models to implicitly learn reaction patterns from large-scale data without requiring explicit chemical knowledge or specialized cheminformatics tools.

\begin{figure}
\centerline{\includegraphics[width=\linewidth]{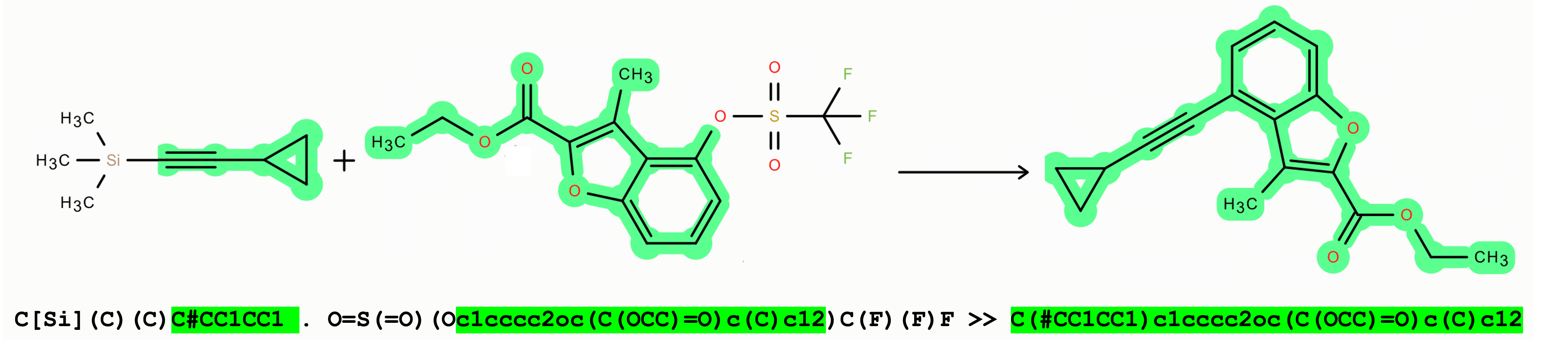}}
\caption{Example of Structural Invariance in a Chemical Reaction. Green highlighting indicates molecular scaffolds that remain unchanged during the reaction, while modified regions represent reactive sites.} 
\label{fig0}
\end{figure}

\begin{figure*}
\centerline{\includegraphics[clip, trim=1.7cm 1.7cm 0cm 0cm, width=\linewidth]{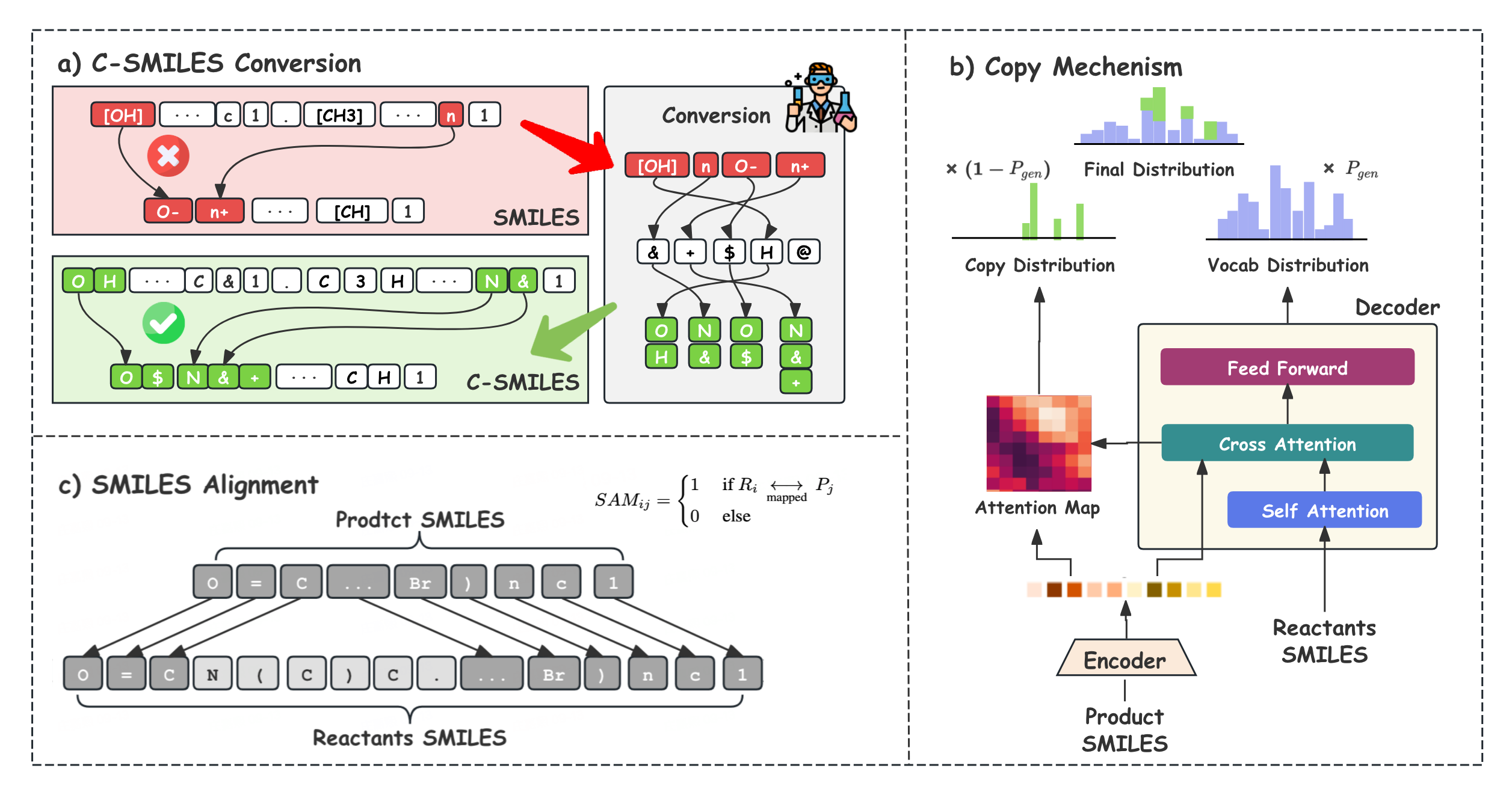}}
\caption{Overview of C-SMILES and Copy Mechanism based on Transformer.} 
\label{fig1}
\end{figure*}

\textbf{Problem}. Despite their promise, current template-free methods face a fundamental limitation: they fail to effectively capture the \textit{structural invariance} inherent in chemical reactions~\cite{editretro}. As illustrated in Figure~\ref{fig0}, most chemical transformations preserve substantial portions of the molecular scaffold while modifying only localized reactive sites~\cite{g2gs}. This structural conservation represents a key principle in organic chemistry, yet existing sequence-based approaches treat molecular generation as unconstrained text generation, leading to unnecessary modifications of chemically stable regions. Consequently, these methods exhibit larger molecular editing distances between reactants and products, expanded search spaces, and reduced chemical validity in predicted outcomes~\cite{rsmiles}. This limitation becomes particularly problematic for complex synthetic routes and rare reaction types, ultimately restricting the practical applicability of template-free approaches in real-world drug discovery scenarios.

To address these challenges, we introduce a novel framework that explicitly models structural invariance in template-free retrosynthesis prediction. Our approach tackles two key objectives: (i) developing a molecular representation that captures atom-level correspondences between reactants and products, and (ii) constraining the generation process to preserve unchanged molecular regions while selectively modifying reactive sites. We propose C-SMILES, a structure-aware molecular representation that decomposes traditional SMILES into element-token pairs using specialized tokens, effectively reducing the sequence editing distance between reactants and products. Building upon this representation, we incorporate a copy-augmented generation mechanism inspired by recent advances in natural language processing, which dynamically determines whether to preserve existing molecular fragments or generate new structural elements. Additionally, we introduce SMILES alignment guidance that leverages chemical atom mappings to enhance attention consistency during the generation process.

Our contributions are summarized as follows:

\begin{itemize}
    \item We propose C-SMILES, a novel molecular representation that captures structural invariance by minimizing editing distance between reactants and products through element-token decomposition with specialized vocabulary.
    \item We introduce a copy-augmented generation framework that preserves unchanged molecular structures while selectively modifying reactive sites, effectively constraining the search space for chemically valid predictions.
    \item We demonstrate state-of-the-art performance on both USPTO-50K and large-scale USPTO-FULL benchmarks, achieving significant improvements in accuracy and validity compared to existing template-free methods.
\end{itemize}

\section{Method}

\subsection{Overview}
Our framework addresses the structural invariance challenge in template-free retrosynthesis through three key innovations: (1) C-SMILES, a novel molecular representation that decomposes traditional SMILES into element-token pairs to minimize editing distance; (2) a copy-augmented generation mechanism that dynamically preserves unchanged molecular fragments while selectively modifying reactive sites; and (3) attention alignment guidance that leverages chemical atom mappings to enhance generation consistency. Figure~\ref{fig1} illustrates the overall architecture of our approach.

\begin{table}
\caption{Description of Special Token in C-SMILES}
\begin{center}
\begin{tabular}{cl}
\hline
\textbf{Special Token} & \multicolumn{1}{c}{\textbf{Description}}                 \\ \hline
\&                     & Lowercase. Example: c -\textgreater C, \&.               \\
+                      & Positive charges. Example: {[}S+{]} -\textgreater S, +.  \\
\$                     & Negative charges. Example: {[}N-{]} -\textgreater N, \$. \\
H                      & Hydrogen atoms. Example: {[}SH{]} -\textgreater S, H.    \\
@                      & Chiral center. Example: {[}S@{]} -\textgreater S, @.     \\ \hline   
\end{tabular}
\end{center}
\label{tab1}
\end{table}

\subsection{C-SMILES Representation}
Traditional SMILES representations often exhibit large editing distances between reactants and products due to subtle atomic state changes that do not reflect the underlying structural invariance. As shown in Figure~\ref{fig1}a, conventional SMILES may represent similar atomic environments with different tokens, hindering model to capture structural conservation.

To address this limitation, we propose C-SMILES, which decomposes complex atomic representations into element-token pairs using five special tokens: [\&, +, \$, H, @]. Table~\ref{tab1} details the functionality of each special token. This decomposition strategy offers two key advantages: (1) it explicitly separates elemental identity from atomic properties, enabling better tracking of invariant elements across reactions; and (2) it reduces the vocabulary size from 80 to 55 tokens by covering 30 complex tokens with simple element-token combinations, thereby constraining the generation search space.

C-SMILES conversion process transforms bracketed atomic representations into decomposed sequences. For example, [OH] becomes \underline{O, H} and [s+] becomes \underline{S, \&, +}. This representation maintains chemical validity while creating more consistent sequence alignments between reactants and products.
% , facilitating the subsequent copy mechanism.

\begin{algorithm}[h]
\caption{Copy-Augmented Generation}
\label{algorithm1}
\begin{algorithmic}[1]
\REQUIRE Product sequence $P = \{p_1, \ldots, p_{|P|}\}$
\ENSURE Reactant sequence $R$
\STATE $H \leftarrow \text{Encoder}(P)$
\STATE Initialize $s_0 \leftarrow \texttt{<SOS>}$, $R \leftarrow \emptyset$
\FOR{$t = 1$ to $T$}
    \STATE Compute cross-attention: $a^t \leftarrow \text{Attention}(s_{t-1}, H)$
    \STATE Compute context: $h_t^* \leftarrow \sum_{i} a_i^t h_i$
    \STATE Compute generation probability: $p_{\text{gen}}^t \leftarrow \sigma(W[h_t^*; s_{t-1}; x_t] + b)$
    \STATE Combine distributions: $P(w) \leftarrow p_{\text{gen}}^t P_{\text{vocab}}(w) + (1-p_{\text{gen}}^t) \sum_{i:p_i=w} a_i^t$
    \STATE Sample token: $r_t \leftarrow \text{Sample}(P(w))$
    \STATE Update: $R \leftarrow R \cup \{r_t\}$, $s_t \leftarrow \text{Decoder}(s_{t-1}, r_t)$
    \IF{$r_t = \texttt{<EOS>}$} \STATE \textbf{break} \ENDIF
\ENDFOR
\RETURN $R$
\end{algorithmic}
\end{algorithm}

\subsection{Copy-Augmented Generation}
Building upon the C-SMILES representation, we introduce a copy-augmented generation framework that dynamically determines whether to preserve existing molecular fragments or generate new structural elements. Inspired by pointer-generator networks in natural language processing~\cite{copy,pointnetwork}, our approach combines the generative capabilities of Transformers with selective copying from the input product.

At each decoding step $t$, we compute a generation probability $p_{\text{gen}}^t$ that controls the balance between copying and generation:

\begin{equation}
p_{\text{gen}}^t = \sigma\left(w_h^T h_t^* + w_s^T s_t + w_x^T x_t + b\right),
\end{equation}

where $h_t^*$ is the attention-weighted context vector computed as:

\begin{equation}
h_t^* = \sum_{i=1}^{|P|} a_i^t h_i,
\end{equation}

and $s_t$, $x_t$ represent the current decoder state and input embedding, respectively. The cross-attention weights $a_i^t$ indicate the relevance of each product token $p_i$ for the current prediction step.

The final token probability distribution combines vocabulary generation and copying distributions:

\begin{equation}
P(w) = p_{\text{gen}}^t P_{\text{vocab}}(w) + (1-p_{\text{gen}}^t) \sum_{i:p_i=w} a_i^t,
\end{equation}

where $P_{\text{vocab}}(w)$ is the standard vocabulary distribution and the second term aggregates attention weights for tokens $w$ that appear in the product sequence.

To enhance the copy mechanism's effectiveness, we introduce a copy index prediction auxiliary task that explicitly supervises the model's copying decisions. The copy index loss is defined as:

\begin{equation}
\mathcal{L}_{CI} = -\sum_{t=1}^T \left(y_t \log p_{\text{gen}}^t + (1-y_t) \log (1-p_{\text{gen}}^t)\right),
\end{equation}

where $y_t$ indicates the ground-truth copying decision at step $t$. During training, we apply teacher forcing~\cite{teacher} with gradual annealing~\cite{annealing} to guide the model toward optimal copying behavior.

\subsection{Attention Alignment Guidance}
To further enhance the copy mechanism's accuracy, we leverage chemical atom mappings to guide the model's attention patterns. Chemical reactions typically preserve atom correspondences, which can be extracted from atom-mapped reaction data to create a SMILES Alignment Map (SAM).

The SAM construction process, detailed in Algorithm~\ref{algorithm2}, maps tokens between reactant and product sequences based on their underlying atom correspondences:

\begin{equation}
\text{SAM}_{ij} = \begin{cases}
1 & \text{if } R_i \leftrightarrow P_j \text{ (atom-mapped)} \\
0 & \text{otherwise}
\end{cases}
\end{equation}

\begin{algorithm}[b]
\caption{SMILES Alignment Map Construction}
\label{algorithm2}
\begin{algorithmic}[1]
\REQUIRE Atom-mapped product $S_P$ and reactants $S_R$
\ENSURE SMILES Alignment Map $\text{SAM}$
\STATE Initialize $\text{SAM} \leftarrow \emptyset$
\STATE Tokenize: $T_P \leftarrow \text{Tokenize}(S_P)$, $T_R \leftarrow \text{Tokenize}(S_R)$
\FOR{each atom-mapped token $r_i \in T_R$}
    \STATE Find corresponding $p_j \in T_P$ with same atom map
    \WHILE{tokens match or share atom mapping}
        \STATE Add alignment: $\text{SAM}_{ij} \leftarrow 1$
        \STATE Advance pointers: $i \leftarrow i+1$, $j \leftarrow j+1$
    \ENDWHILE
\ENDFOR
\STATE Extend alignments to adjacent non-atom tokens
\RETURN $\text{SAM}$
\end{algorithmic}
\end{algorithm}

Inspired by guided attention approaches~\cite{guiding}, we encourage attention consistency with the ground-truth alignment through a guidance loss. To improve robustness, we apply label smoothing~\cite{labelsmoothing} to the binary alignment labels:

\begin{equation}
\tilde{\text{SAM}}_{ij} = (1-\epsilon) \cdot \text{SAM}_{ij} + \epsilon \cdot u_{ij},
\end{equation}

where $\epsilon$ is the smoothing factor and $u_{ij}$ represents a uniform distribution. The alignment loss is then computed as:
{\small
\begin{equation}
\mathcal{L}_{SA} = -\frac{1}{T} \sum_{t=1}^T \sum_{j=1}^{|P|} \left( \tilde{\text{SAM}}_{ij} \log a_j^t + (1 - \tilde{\text{SAM}}_{ij}) \log (1 - a_j^t) \right),
\end{equation}
}

This guidance mechanism enhances both the interpretability and effectiveness of the copy mechanism by aligning model attention with chemical knowledge.

\begin{table*}[t]
\centering
\caption{Top-k accuracy on USPTO-50K. Best performance is in \textbf{bold}.}
\label{tab2}
\begin{tabular}{lccccccccc}
\toprule
\multicolumn{1}{c}{\multirow{4}{*}{\textbf{Model}}}
    & \multicolumn{9}{c}{\textbf{Top-k accuracy (\%)}} \\ 
    \cmidrule(lr){2-10}
    & \multicolumn{4}{c}{\textbf{Reaction class known}} 
    & & \multicolumn{4}{c}{\textbf{Reaction class unknown}} \\ 
    \cmidrule(lr){2-5} \cmidrule(lr){7-10}
    & Top-1 & Top-3 & Top-5 & Top-10 
    & & Top-1 & Top-3 & Top-5 & Top-10 \\
\midrule
\textit{\textbf{\# Template-Based}} & & & & & & & & & \\
RetroSim         & 52.9 & 73.8 & 81.2 & 88.1 && 37.3 & 54.7 & 63.3 & 74.1 \\
GLN              & \textbf{64.2} & 79.1 & 85.2 & 90.0 
                 && 52.5 & 69.0 & 75.6 & 83.7 \\
LocalRetro        & 63.9 & \textbf{86.8} & \textbf{92.4} & \textbf{96.3} && \textbf{53.4} & \textbf{77.5} & \textbf{85.9} & \textbf{92.4} \\
\midrule
\textit{\textbf{\# Semi-Template}} & & & & & & & & & \\
RetroXpert       & 62.1 & 75.8 & 78.5 & 80.9 && 50.4 & 61.1 & 62.3 & 63.4 \\
G2G              & 61.0 & 81.3 & 86.0 & 88.7 && 48.9 & 67.6 & 72.5 & 75.1 \\
GraphRetro       & 63.9 & 81.5 & 85.2 & 88.1 && 53.7 & 68.3 & 72.2 & 75.5 \\
RetroPrime       & 64.8 & 81.6 & 85.0 & 86.9 && 51.4 & 70.8 & 74.0 & 75.5 \\
MEGAN            & 60.7 & 82.0 & 87.5 & 91.6 && 48.1 & 70.7 & 78.4 & 86.1 \\
Graph2Edits      & \textbf{67.1} & \textbf{87.5} & \textbf{91.5} & \textbf{93.8} && \textbf{55.1} & \textbf{77.3} & \textbf{83.4} & \textbf{89.4} \\
\midrule
\textit{\textbf{\# Template-Free}} & & & & & & & & & \\
SCROP            & 59.0 & 74.8 & 78.1 & 81.1 && 43.7 & 60.0 & 65.2 & 68.7 \\
Aug. Transformer &   -  &   -  &   -  &   -  && 48.3 &   -  & 73.4 & 77.4 \\
Graph2SMILES     &   -  &   -  &   -  &   -  && 52.9 & 66.5 & 70.0 & 72.9 \\
GTA              & - & - & - & - && 51.1 & 67.6 & 74.8 & 81.6 \\
Retroformer      & 64.0 & 82.5 & 86.7 & 90.2 && 53.2 & 71.1 & 76.6 & 82.1 \\  \rowcolor{gray!20}
\textbf{C-SMILES (Ours)} 
                 & \textbf{67.2} & \textbf{87.9} & \textbf{91.5} & \textbf{94.4} 
                 && \textbf{55.2} & \textbf{77.6} & \textbf{83.9} & \textbf{89.1} \\
\bottomrule
\end{tabular}
\end{table*}

\subsection{Training Strategy}
Our complete training objective combines three complementary loss functions:

\begin{equation}
\mathcal{L} = \mathcal{L}_{LM} + \lambda_{SA} \mathcal{L}_{SA} + \lambda_{CI} \mathcal{L}_{CI},
\end{equation}

where $\mathcal{L}_{LM}$ is the standard language modeling loss (implemented as R-Drop~\cite{rdrop}), $\mathcal{L}_{SA}$ is the attention alignment loss, and $\mathcal{L}_{CI}$ is the copy index prediction loss. The hyperparameters $\lambda_{SA}$ and $\lambda_{CI}$ balance the relative contributions of each component. During training, we gradually reduce teacher forcing for the copy mechanism to encourage autonomous learning of copying strategies.

\section{Experiments}

\subsection{Experimental Setup}

\subsubsection{Datasets}
We evaluate our method on two widely-used retrosynthesis prediction benchmarks to demonstrate its effectiveness across different scales. \textbf{USPTO-50K}~\cite{uspto50k} is a standard benchmark containing 50,016 atom-mapped reactions extracted from the USPTO patent database. Following the established protocol in~\cite{retrosim}, we use the same data split consisting of 40,008 reactions for training, 5,001 for validation, and 5,007 for testing. \textbf{USPTO-FULL}~\cite{gln} represents a larger-scale evaluation benchmark with approximately 1,004,118 atom-mapped reactions from the USPTO database. After removing duplicates and erroneous entries following standard preprocessing protocols~\cite{gln}, the dataset contains 801,496 training reactions, 101,311 validation reactions, and 101,311 test reactions. For both datasets, we apply root-aligned SMILES augmentation~\cite{rsmiles} on-the-fly during training to improve model robustness, and convert all molecular representations to our proposed C-SMILES format before training and evaluation.

\subsubsection{Evaluation Metrics}
We adopt three complementary evaluation metrics to comprehensively assess model performance. \textbf{Top-k Accuracy} measures the percentage of test cases where the ground-truth reactants appear among the top-k predictions generated by beam search. \textbf{Top-k Validity} assesses the chemical validity of generated predictions by verifying whether the top-k predicted reactant SMILES can be successfully parsed by {\tt RDKit}~\cite{rdkit}. \textbf{Top-k Round-Trip Accuracy} evaluates the chemical plausibility of predictions by measuring the percentage of predicted reactants that can successfully reproduce the original product molecule using the Molecular Transformer~\cite{molecular_transformer} as the forward reaction prediction model. All evaluations use beam search decoding~\cite{beamsearch} with beam size 10, and we report results for $k \in \{1, 3, 5, 10\}$ following standard practice in the field.

\subsubsection{Baselines}
We compare our approach against representative methods from three categories of retrosynthesis prediction. Template-based methods include RetroSim~\cite{retrosim} , GLN~\cite{gln} and LocalRetro~\cite{localretro}. Semi-template methods include RetroXpert~\cite{xpert}, G2Gs~\cite{g2gs}, GraphRetro~\cite{graphretro}, RetroPrime~\cite{retroprime}, MEGAN~\cite{megan} and Graph2Edits~\cite{graph2edit}. Template-free methods include SCROP~\cite{scrop}, Augmented Transformer~\cite{augtrans}, Graph2SMILES~\cite{graph2smiles}, GTA~\cite{gta} and Retroformer~\cite{retroformer}. We exclude large-scale pre-trained models (e.g. ChemFormer~\cite{chemformer}, EditRetro~\cite{editretro}) from our comparison as they leverage additional training data beyond our experimental setup and may involve potential data leakage issues as discussed in~\cite{editretro}. All baseline results are taken from their original publications.

\subsubsection{Implementation Details}
Our model is implemented using PyTorch and trained on NVIDIA RTX 4090 GPUs. We employ the Adam optimizer~\cite{adam} with $\beta_1 = 0.9$, $\beta_2 = 0.98$, $\epsilon = 1 \times 10^{-9}$, initial learning rate of $1 \times 10^{-4}$, and batch size of 32. The model uses a 6-layer Transformer encoder-decoder architecture with 8 attention heads and hidden dimension of 512. Key hyperparameters include attention alignment loss weight $\lambda_{SA} = 0.1$ and copy index loss weight $\lambda_{CI} = 0.1$, which are tuned on the validation set. The label smoothing factor $\epsilon$ is set to 0.1 for the alignment guidance, and teacher forcing annealing follows a linear schedule from 1.0 to 0.1 over the first 50\% of training epochs. Code are available at \underline{https://anonymous.4open.science/r/C-SMILES-0484}.

\subsection{Performance On USPTO-50K}

\subsubsection{Top-k Accuracy}
As shown in Table~\ref{tab2}, our method utilizing C-SMILES achieves state-of-the-art performance across both known and unknown reaction types among template-free methods. Specifically, C-SMILES achieves 67.2\% top-1 accuracy under known reaction conditions and 55.2\% under unknown conditions, representing substantial improvements over the strongest template-free baseline Retroformer (64.0\% and 53.2\%, respectively). This corresponds to a 5.0\% relative improvement under known conditions and a 3.8\% relative improvement under unknown conditions. Notably, our template-free approach not only outperforms all existing template-free methods but also achieves competitive performance compared to template-based and semi-template approaches that rely on predefined chemical knowledge. The consistent improvements across all top-k metrics demonstrate the effectiveness of our structural invariance preservation mechanism in constraining the search space while maintaining high prediction accuracy.

\begin{table}[t]
\caption{Top-k Validity on USPTO-50K with reaction class unknown. Best performance is in \textbf{bold}.}
\begin{center}
\begin{tabular}{lcccc}
\toprule
\multicolumn{1}{c}{\multirow{2.5}{*}{\textbf{Model}}} & \multicolumn{4}{c}{\textbf{Top-k Validity (\%)}}  \\  \cmidrule{2-5}
                                    & 1             & 3             & 5             & 10                \\  \midrule
                                    
Graph2SMILES                        & 99.4          & 90.9          & 84.9          & 74.9              \\
RetroPrime                          & 98.9          & 98.2          & 97.1          & 92.5              \\
Retroformer                         & 99.2          & 98.5          & 97.4          & 96.7              \\ \rowcolor{gray!20}
\textbf{C-SMILES (Ours)}            & \textbf{99.9} & \textbf{99.4} & \textbf{98.8} & \textbf{97.0}     \\
\bottomrule
\end{tabular}
\label{tab3}
\end{center}
\end{table}

\subsubsection{Top-k Validity}
Table~\ref{tab3} presents the molecular validity results comparing our approach with representative template-free methods. We focus on template-free methods for this evaluation since template-based and semi-template methods inherently ensure validity through their use of predefined templates and chemical editing tools, making validity comparison less meaningful. Our model demonstrates superior performance in generating chemically valid SMILES, achieving 99.9\% top-1 validity and maintaining 97.0\% validity even at top-10 predictions. This represents a notable improvement over existing template-free approaches, particularly Graph2SMILES which shows degradation in validity for higher-k predictions. The high validity rates indicate that our copy-augmented generation framework effectively preserves chemically meaningful structures while exploring the molecular space.

\subsubsection{Top-k Round-Trip Accuracy}
To evaluate the chemical plausibility of our predictions, we assess round-trip accuracy using the Molecular Transformer~\cite{moltrans} as the forward reaction model. As shown in Table~\ref{tab4}, C-SMILES achieves the highest round-trip accuracy across all metrics, with 90.0\% top-1 round-trip accuracy compared to the best baseline Retroformer at 78.9\%. This 11.1\% improvement demonstrates that our predicted reactants are not only syntactically valid but also chemically plausible precursors capable of reproducing the original product through forward synthesis. The substantial improvement in round-trip consistency can be attributed to our structural invariance mechanism, which guides the model to generate chemically coherent transformations that respect the underlying reaction chemistry.

\begin{table}[t]
\caption{Top-k Round-Trip Accuracy on USPTO-50K with reaction class unknown. Best performance is in \textbf{bold}.}
\begin{center}
\begin{tabular}{lcccc}
\toprule
\multicolumn{1}{c}{\multirow{2.5}{*}{\textbf{Model}}} & \multicolumn{4}{c}{\textbf{Top-k Round-Trip Accuracy (\%)}}  \\  \cmidrule{2-5}
                                    & 1             & 3             & 5             & 10                \\  \midrule
                                    
Graph2SMILES                        & 76.7          & 56.0          & 46.6          & 34.9              \\
RetroPrime                          & 79.6          & 59.6          & 50.3          & 40.4              \\
Retroformer                         & 78.9          & 72.0          & 67.1          & 57.2              \\ \rowcolor{gray!20}
\textbf{C-SMILES (Ours)}            & \textbf{90.0} & \textbf{80.3} & \textbf{73.8} & \textbf{67.8}     \\
\bottomrule
\end{tabular}
\label{tab4}
\end{center}
\end{table}

\begin{table}[b]
\centering
\caption{Top-k accuracy for retrosynthesis prediction on USPTO-FULL without reaction class. The metric values of other models are taken from their original papers. Best performance is in \textbf{bold}.}
\label{tab5}
\begin{tabular}{lcccc}
\toprule
\multicolumn{1}{c}{\multirow{2.5}{*}{\textbf{Model}}} & \multicolumn{4}{c}{\textbf{Top-k Accuracy (\%)}}                  \\  \cmidrule{2-5}
\multicolumn{1}{c}{}                                & 1             & 3             & 5             & 10                \\  \midrule
\textbf{Template-Based}                             &               &               &               &                   \\  \midrule
RetroSim                                            & 32.8          & -             & -             & 56.1              \\
GLN                                                 & 39.3          & -             & -             & 63.7              \\
LocalRetro                                          & 39.1          & 53.3             & 58.4             & 63.7              \\  \midrule
\textbf{Semi-Template-Based}                        &               &               &               &                   \\  \midrule
RetroPrime                                          & 44.1          & 59.1          & 62.8          & 68.5              \\ 
MEGAN                                               & 33.6          & -             & -             & 63.9              \\
Graph2Edits                                         & 44.0          & 60.9          & 66.8          & 72.5              \\  \midrule 
\textbf{Template-Free}                              &               &               &               &                   \\  \midrule
Aug. Transformer                                    & 46.2          & -             & -             & 73.3              \\
GTA                                                 & 46.6          & -             & -             & 70.4              \\
Graph2SMILES                                        & 45.7          & -             & -             & 63.4              \\ \rowcolor{gray!20}
\textbf{C-SMILES (Ours)}                            & \textbf{48.9} & \textbf{64.2} & \textbf{70.5} & \textbf{75.8}     \\
\bottomrule
\end{tabular}
\end{table}

\subsection{Performance On USPTO-FULL}

To assess the scalability of our approach, we conduct experiments on the large-scale USPTO-FULL dataset, which contains over 1 million atom-mapped reactions and represents a more comprehensive and challenging evaluation benchmark compared to USPTO-50K. Table~\ref{tab5} presents the comparative results on this large-scale dataset.

C-SMILES achieves 50.8\% top-1 accuracy, outperforming the template-free baseline Augmented Transformer by 4.6\%. Our method also demonstrates superior performance across all top-k metrics, achieving 67.5\% top-3 accuracy and 78.0\% top-10 accuracy. Notably, C-SMILES surpasses not only template-free methods but also semi-template approaches such as Graph2Edits (44.0\% top-1) and RetroPrime (44.1\% top-1), despite not relying on predefined reaction templates or chemical editing tools.

The 20-fold dataset increase demonstrates our method's scalability, with C-SMILES maintaining consistent improvements despite massive scale expansion. 
The 6.7\% improvement over Graph2Edits is particularly significant given its sophisticated graph editing operations. 
% Our superior top-10 performance indicates effective handling of increased search space complexity, while the improvement from top-1 to top-10 accuracy shows our copy-augmented generation consistently ranks plausible alternatives highly.

The consistent performance gains across this large-scale dataset demonstrate that our structural invariance preservation mechanism scales effectively to diverse real-world chemical reactions. The ability to maintain high accuracy on such a comprehensive dataset underscores the practical applicability of our approach for industrial-scale retrosynthesis prediction.

\begin{table}[b]
\setlength{\tabcolsep}{4pt}
\caption{ablation study of C-SMILES components with reaction class unknown on USPTO-50K dataset. (a) Copy mechanism, (b) SMILES alignment guidance, (c) Copy index prediction. \\ Best performance is in \textbf{bold}.}
\label{tab6}
\centering
\begin{tabular}{ccccccccccccc}
\toprule
& \multirow{2.5}{*}{\textbf{(a)}} 
& \multirow{2.5}{*}{\textbf{(b)}} 
& \multirow{2.5}{*}{\textbf{(c)}} 
& \multicolumn{4}{c}{\textbf{Top-k Accuracy (\%)}} 
& \multicolumn{4}{c}{\textbf{Top-k Validity (\%)}} \\
\cmidrule(lr){5-8} \cmidrule(lr){9-12}
& & & 
& 1 & 3 & 5 & 10 
& 1 & 3 & 5 & 10 \\
\midrule
\ding{172} 
& - & $\checkmark$ & $\checkmark$ 
& 53.0 & 72.8 & 78.2 & 81.8 
& 99.1 & 98.3 & 97.3 & 95.6 \\
\ding{173} 
& $\checkmark$ & - & $\checkmark$ 
& 53.8 & 71.9 & 79.0 & 85.6 
& 99.5 & 99.2 & 97.7 & 95.3 \\
\ding{174} 
& $\checkmark$ & $\checkmark$ & - 
& 54.3 & 75.8 & 82.8 & 88.8 
& 99.8 & 99.3 & \textbf{98.8} & 96.2 \\
\ding{175} 
& $\checkmark$ & $\checkmark$ & $\checkmark$ 
& \textbf{55.2} & \textbf{77.6} & \textbf{83.9} & \textbf{89.1} 
& \textbf{99.9} & \textbf{99.4} & \textbf{98.8} & \textbf{97.0} \\
\bottomrule
\end{tabular}
\end{table}

\subsection{Ablation Studies}

To investigate the individual contributions of our key components, we conduct ablation studies on the USPTO-50K dataset under unknown reaction conditions. We evaluate three main components: (a) Copy mechanism, (b) SMILES alignment guidance, and (c) Copy index prediction.

As shown in Table~\ref{tab6}, all three components contribute positively to the final performance. To analyze the individual contribution of the copy mechanism, we compare the model without it \ding{172} against the full model \ding{175}. The copy mechanism contributes a 2.2\% improvement in top-1 accuracy (from 53.0\% to 55.2\%) and substantial gains at higher k values, with top-10 accuracy improving by 7.3\% (from 81.8\% to 89.1\%). This demonstrates the copy mechanism's effectiveness in both accuracy and prediction diversity.

For SMILES alignment guidance, comparing \ding{173} (without alignment) to \ding{175} (full model) shows its significant contribution, with top-1 accuracy improving by 1.4\% (from 53.8\% to 55.2\%) and top-3 accuracy by 5.7\% (from 71.9\% to 77.6\%). The validity scores also improve notably, indicating that alignment guidance helps generate more chemically coherent structures by leveraging atom mapping information.

Copy index prediction contributes additional refinement, as seen by comparing \ding{174} (without copy index prediction) to \ding{175} (full model). This component provides a 0.9\% improvement in top-1 accuracy (from 54.3\% to 55.2\%) and consistent gains across all metrics. The improvement demonstrates the value of explicit supervision for the model's copying decisions.

The full model \ding{175} achieves the best performance across both accuracy and validity metrics, with all components working synergistically. The consistent improvements in validity scores indicate that our components collectively enhance both prediction accuracy and chemical validity.

\begin{figure}
\centerline{\includegraphics[clip, trim=2.5cm 0cm 1.5cm 0cm, width=\linewidth]{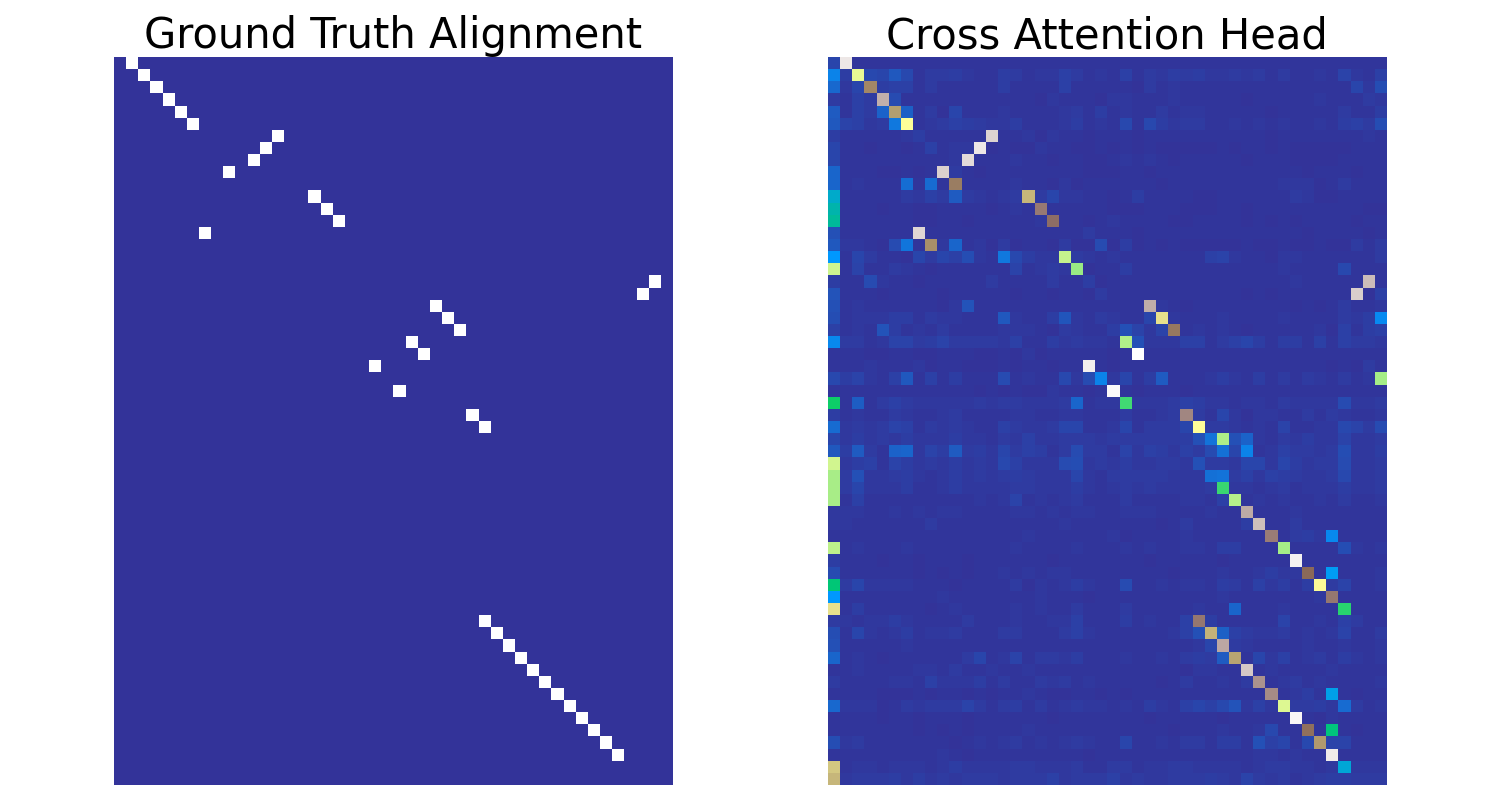}}
\caption{SMILES Alignment (left) and Cross-Attention score (right).} 
\label{fig2}
\end{figure}

\subsection{Attention Map}
Figure~\ref{fig2} shows the SMILES alignment between products and reactants (left) and the cross-attention score from the first attention head in final decoder layer (right). It is evident that after SMILES Alignment Loss training, the cross-attention computation effectively captures the corresponding information from the SMILES alignment map. This not only assists the model generation process during the decoding phase but also provides a strong reference for the copy mechanism, significantly improving the accuracy of predicting copy token and preserving structural invariance, thereby enhancing overall retrosynthesis performance.

\subsection{Case Study}
\begin{figure}[b]
\centerline{\includegraphics[clip, trim=0cm 2.0cm 0cm 0cm, width=1.0\linewidth]{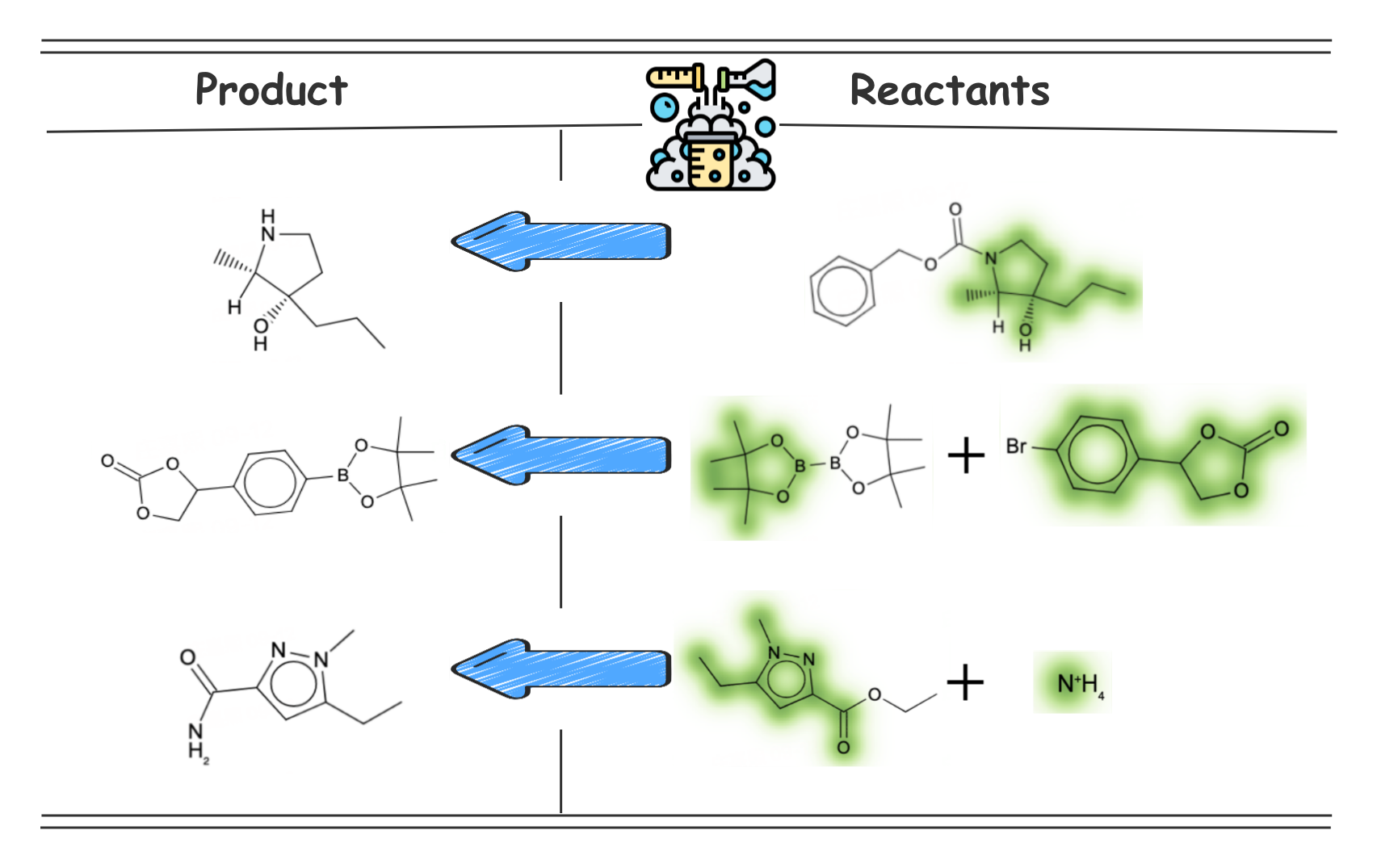}}
\caption{Retrosynthesis Prediction under Copy Mechanism. Highlight in green when atoms/bonds are directly copied from Products.} 
\label{fig3}
\end{figure}

Figure~\ref{fig3} shows the cases inferred by our method. In these cases, atoms/bonds with copy token probability greater than the 0.7 threshold are highlighted in green. As shown in the figure, our model is able to infer copy index with high confidence (\textgreater 0.7) and directly copy the corresponding tokens from the product based on these index. Compared to a full vocabulary search, our method significantly reduces the model’s search space, improving efficiency while maintaining accuracy and validity.

\section{Component Synergy Analysis}

Our framework exhibits a notable synergistic relationship between the copy mechanism and SMILES alignment guidance, where both components leverage the decoder's cross-attention mechanism in complementary ways. The copy mechanism relies on cross-attention weights to determine which product tokens should be copied versus generated, using these attention scores to construct the copying probability distribution in Equation (3). Simultaneously, the SMILES alignment module introduces chemical prior knowledge through atom mappings, explicitly guiding the cross-attention patterns to align with ground-truth molecular correspondences via the alignment loss in Equation (7).

This complementary design creates a mutually reinforcing system where each component enhances the effectiveness of the other. The SMILES alignment guidance improves the quality of cross-attention weights by encouraging chemically meaningful attention patterns, which in turn provides more accurate copying decisions for copy mechanism. Conversely, copy mechanism's selective preservation of molecular fragments reduces the burden on attention alignment task by maintaining structural consistency throughout generation process.

The synergy between these components manifests in two key benefits. First, it enhances prediction accuracy by ensuring that copying decisions are informed by chemically grounded attention patterns rather than purely learned associations. Second, it improves model interpretability, as the guided attention explicitly captures atomic correspondences that align with chemical intuition. This synergistic design represents a principled approach to incorporating chemical knowledge into neural generation models while maintaining the flexibility of end-to-end learning.
\section{Related work}
\textbf{Template-based methods} leverage reaction templates from curated databases to encode the rules of chemical transformations and match input products with appropriate templates. RetroSim \cite{retrosim} adopts a similarity-driven strategy by ranking candidate templates based on molecular fingerprint similarity. GLN \cite{gln} employs a Conditional Graph Logic Network to determine when specific template rules should be applied, implicitly incorporating chemical feasibility and synthetic strategy. Building on the chemical intuition that reactions are often governed by local structural changes, LocalRetro \cite{localretro} encodes local environments while allowing refinement to accommodate non-local transformations. Although template-based methods offer strong interpretability, their dependence on predefined templates constrains both coverage and scalability, limiting their applicability to complex real-world scenarios.

\textbf{Semi-template methods} relax the strict reliance on templates by combining chemical heuristics with generative models. Most approaches adopt a two-stage framework: (i) \textit{reaction center identification}, which detects the reactive site of the product and cleaves bonds to generate intermediate structures known as synthons; and (ii) \textit{synthons completion}, which reconstructs complete reactants from synthons. RetroXpert \cite{xpert} introduces an Edge-enhanced Graph Attention Network (EGAT) to predict bond disconnection probabilities on molecular graphs and a subsequent seq2seq model to generate reactant SMILES. G2Gs \cite{g2gs}, in contrast, performs synthon completion directly in graph space via node selection and edge labeling. GraphRetro \cite{graphretro} extends synthons by attaching leaving groups, while RetroPrime \cite{retroprime} incorporates auxiliary SMILES labels to improve the two-stage pipeline. MEGAN \cite{megan} and Graph2Edits \cite{graph2edit} reformulate the generation process as a sequence of graph edit operations executed with {\tt RDKit}.

\textbf{Template-free methods} cast retrosynthesis prediction (product$\rightarrow$reactants) as a machine translation task (source$\rightarrow$target) \cite{machinetranslation}, enabling the use of advanced NLP architectures such as LSTMs \cite{lstm} and Transformers \cite{transformer}. SCROP \cite{scrop} augments this paradigm with a self-correcting transformer to mitigate syntax errors, while Augmented Transformer \cite{augtrans} improves robustness through SMILES permutation data augmentation. To achieve permutation invariance, Graph2SMILES \cite{graph2smiles} replaces the sequential encoder with a graph-based encoder, and GTA \cite{gta} integrates graph structure via adjacency-matrix-masked self-attention. Retroformer \cite{retroformer} further introduces a local attention mechanism that facilitates information exchange between reactive substructures and the global molecular context.Despite their scalability and automation advantages, template-free methods often lack explicit modeling of structural invariance between products and reactants. In chemical reactions, large portions of the molecular scaffold typically remain unchanged, yet conventional sequence-based models tend to over-modify these invariant regions. This results in larger molecular editing distances, reduced chemical validity, and expanded irrelevant search space, ultimately degrading prediction accuracy and generalization, particularly for rare or complex reactions.
% We propose Retro3D, an end-to-end Transformer that integrates 3D conformer for Retrosynthesis. 
% With the proposed Atom-align Fusion module, we can integrate token and 3D position information ensuring the alignment between them. Next, we propose Distance-weighted Attention mechanism to guide and refine the redistribution of self-attention calculation with spatial distances and refine it across multiple layers. 
% % We also utilize SMILES alignment and two types of data augmentation to enhance our model. 
% % We validate our model on the USPTO-50K dataset with top-k accuracy and validity, demonstrating that Retro3D 
% Our model achieves a new state-of-the-art in template-free methods and is also highly competitive with template-based and semi-template methods. 
% % In future work, we will attempt to introduce our method into multi-step synthesis route planning and explore potential of 3D conformer in Retrosynthesis.
\section{Conclusion}

We introduce a novel framework for template-free retrosynthesis prediction that addresses the fundamental challenge of structural invariance in chemical reactions. Our approach combines C-SMILES, a molecular representation that decomposes SMILES into element-token pairs to minimize editing distance, with a copy-augmented generation mechanism that dynamically preserves unchanged molecular fragments while modifying reactive sites.

Extensive experiments on USPTO-50K and USPTO-FULL datasets demonstrate that our method achieves state-of-the-art performance among template-free approaches, with significant improvements in accuracy, validity, and round-trip consistency. The synergistic design of copy mechanism and SMILES alignment guidance provides both enhanced prediction capability and improved interpretability. 

This work establishes a principled approach for incorporating structural invariance into retrosynthesis, offering valuable insights for future developments in computer aided drug design. The demonstrated ability to preserve chemically meaningful structures while exploring synthetic pathways opens new avenues for synthesis planning and drug discovery.

\bibliographystyle{ieeetr}
\bibliography{c-smiles}

\end{document}